\pdfoutput=1
\documentclass[11pt]{article}

\usepackage[]{acl}

\usepackage{times}
\usepackage{latexsym}
\usepackage{amsfonts}
\usepackage{amsmath}
\usepackage{subcaption}
\usepackage{graphicx}
\usepackage{pgfplots} % 画折线图
\usepackage{scalefnt} %用于设置字体

\usepackage{graphicx}
\usepackage{xcolor}
\usepackage{CJKutf8}
\usepackage{tabularx}
\usepackage{booktabs}
\usepackage{makecell}
\usepackage[misc]{ifsym}
\usepackage{amssymb}% http://ctan.org/pkg/amssymb
\usepackage{pifont}% http://ctan.org/pkg/pifont
\usepackage[export]{adjustbox}

\newcommand{\xmark}{\ding{55}}%

\newcommand{\zhsmall}[1]{\begin{CJK*}{UTF8}{gbsn}\small{#1}\end{CJK*}}

\definecolor{color1}{RGB}{199,47,47}
\definecolor{color2}{RGB}{242,129,29}
\definecolor{color3}{RGB}{52,157,53}
\definecolor{color4}{RGB}{142,111,173}

\newcommand{\colortwo}[1]{\textcolor{color2!70!black}{#1}}
\newcommand{\colorthr}[1]{\textcolor{color3!70!black}{#1}}
\newcommand{\colorfou}[1]{\textcolor{color4!70!black}{#1}}

\usepackage[T1]{fontenc}
\usepackage[utf8]{inputenc}

\usepackage{microtype}

\usepackage{inconsolata}

\usepackage{multirow}
\usepackage{ulem}

\title{MC$^{\text{2}}$: Towards Transparent and Culturally-Aware NLP\\for Minority Languages in China}

\author{Chen Zhang\thanks{Equal contribution.},\ \ Mingxu Tao$^*$,\ \  Quzhe Huang$^*$, \\
\textbf{Jiuheng Lin$^*$,\ \  Zhibin Chen,\ \  Yansong Feng\thanks{Corresponding author.}} \\
Peking University \\
{\tt \{zhangch,thomastao,huangquzhe,czb-peking,fengyansong\}@pku.edu.cn} \\
{\tt linjiuheng@stu.pku.edu.cn}}
\begin{document}
\maketitle
\begin{abstract}
Current large language models demonstrate deficiencies in understanding low-resource languages, particularly the minority languages in China.
This limitation stems from the scarcity of available pre-training data.
To address this accessibility challenge, 
we present MC$^\text{2}$, a \textbf{M}ultilingual \textbf{C}orpus of \textbf{M}inority Languages in \textbf{C}hina, which is the largest open-source corpus of its kind so far.
MC$^\text{2}$ includes four underrepresented languages: Tibetan, Uyghur, Kazakh, and Mongolian.
Notably, we focus on the less common writing systems of Kazakh and Mongolian, i.e., Kazakh Arabic script and traditional Mongolian script, respectively, which have been long neglected in previous corpus construction efforts.
Recognizing the prevalence of language contamination within existing corpora, we adopt a quality-centric solution for collecting MC$^\text{2}$, prioritizing accuracy while enhancing diversity.
Furthermore, we underscore the importance of attending to the multiplicity of writing systems, which is closely related to the cultural awareness of the resulting models.
The MC$^\text{2}$ corpus and related models are made public to the community\footnote{\url{https://github.com/luciusssss/mc2_corpus}}.
\end{abstract}

\section{Introduction}

Recently, the rapid development of large language models (LLMs) has been fueled by the availability of high-quality pre-training data. 
However, only a handful of high-resource languages, such as English and Chinese, have benefitted from the advantages of this progress in LLMs.
Despite having a substantial user base, many languages remain excluded from the benefits of these advancements due to a lack of suitable corpora. 
In this paper, we focus on such underrepresented languages, particularly addressing the minority languages in China, including Tibetan, Uyghur, Kazakh, and Mongolian.

\begin{figure}[t] %插入图片
\centering %图片居中
\resizebox{1\columnwidth}{!}{  %用于修改图片大小
    \begin{tikzpicture}
\begin{axis}[
    sharp plot, %控制线的风格
    %title=line chart,%图像标题
    %xmode=normal,% 控制坐标轴为线性
%		ymode=log,% 控制坐标轴为对数
    xmode=log,
    ymode=log,
    ylabel={Tokens in CulturaX (Million)}, %x坐标名
    xlabel={Population (Million)}, %y坐标名
    width=10cm, height=7.5cm,  %设置长和宽
    xmin=1,xmax=2000,  % 设置x坐标范围
    ymin=10., ymax=5000000,  % 设置y坐标范围
    xtick={10,100,1000}, %指定x轴刻度值。如果为空，则自动设置刻度线。即分割坐标轴
    ytick={10,100,1000,10000,100000,1000000}, %指定y轴刻度值。如果为空，则自动设置刻度线。即分割坐标轴
    yticklabels={$0$,$10^2$,$10^3$,$10^4$,$10^5$,$10^6$},
    xlabel near ticks, % 设置x坐标名位置靠近折线图
    ylabel near ticks, % 设置y坐标名位置靠近折线图
    ymajorgrids=true, % 启用/禁用 [公式] 轴上刻度线位置上的网格线
    grid style=dashed, % 设置网格线格式
    %legend style={at={(0.675,0.1)},anchor=south}, % 设置标签位置
    %legend columns=2, %设置标签列数
    separate axis lines,
    %x axis line style= { draw opacity=0 }
    y axis line style= { draw opacity=0 }
 ]

% 使用人口和culturax数据集中的数据量https://huggingface.co/datasets/uonlp/CulturaX
% 语言    人口          数量(文件大小)
% bo    6 million      718M          54,185,000
% ug    13 million     338M          77,677,306
% kk (Arabic)   1.6 million 0M       00000000
% kk (Cyrillic)   13 million  18G    2,802,485,195
% mn (Cyrillic)  2.7 million   12G   1,850,667,656
% mn (Traditional) 6 million  0      00000000

% en 12.68亿 2,846,970,578,793
% ru 2.58亿  737,201,800,363
% es 5.379yi 373,845,662,394
% de 1.316yi 357,030,348,021
% fr 2.766yi 319,332,674,695
% zh 1310 227,055,380,882

\addplot+[only marks,mark=triangle*,mark options={scale=2}, color=color3] plot coordinates { 
        (5.9, 10) (1.6, 10) (11.624257, 77.677) (6.3, 54.185)
}; 

\addplot+[only marks,mark=square*,mark options={scale=1.5}, color=color4] plot coordinates { 
        (1268, 2846970.578) (258, 737201.800) (538, 373845.662) (132,357030.348)
        (277, 318332.674) (1310, 227055.380)
}; 

% Tamil 35m, 4,378,078,610
% Urdu 104m, 2,703,052,627
% Bangla  228m, 9,572,929,804
% Hindi 570m, 16,791,362,871
% Indonesian 170m, 12,062,966,061
% Hebrew 9m, 4,937,152,096
\addplot+[only marks,mark=*,mark options={scale=1.5}, color=color2] plot coordinates { 
        (2.7, 1850.667) (13, 2802.485) (35,4378.078) (104, 2703.052) (228, 9572.929)
        (570, 16791.362) (170, 12062.966) (9, 4937.152)
}; 

%store coodinates
\path[-] (rel axis cs:0,0)     coordinate(botstart)
          --(rel axis cs:0,0.0)coordinate(interruptbotA)
         (rel axis cs:0,0.1)  coordinate(interruptbotB)
         --(rel axis cs:0,1)   coordinate(botstop);

\path[-] (rel axis cs:1,0)     coordinate(topstart)
         --(rel axis cs:1,0.0) coordinate(interrupttopA)
         (rel axis cs:1,0.1)  coordinate(interrupttopB)
         --(rel axis cs:1,1)   coordinate(topstop);
\end{axis}

%Draw the axis with a decoration:
\draw(botstart)-- (interruptbotA) decorate[decoration=zigzag]{--(interruptbotB)} -- (botstop);
\draw(topstart)-- (interrupttopA) decorate[decoration=zigzag]{--(interrupttopB)} -- (topstop);

\node (a0) at (7.95, 4.85){\texttt{zh}};
\node (a1) at (7.91, 5.39){\texttt{en}};
\node (a2) at (6.12, 5.33){\texttt{ru}};
\node (a3) at (6.23, 4.35){\texttt{fr}};
\node (a4) at (6.95, 5.02){\texttt{es}};
\node (a5) at (5.4, 5.02){\texttt{de}};

\node (b0) at (0.7, 0.31){\texttt{kk-Arab}};
\node (b1) at (2.25, 0.28){\texttt{mn-Mong}};
\node (b2) at (2.03, 1.08){\texttt{bo}};
\node (b3) at (2.72, 1.25){\texttt{ug}};

\node (c0) at (1.09, 2.65){\texttt{mn-Cyrl}};
\node (c1) at (2.83, 2.22){\texttt{kk-Cyrl}};
\node (c2) at (2.42, 3.1){\texttt{he}};
\node (c3) at (3.92, 3.05){\texttt{ta}};
\node (c4) at (5.13, 2.78){\texttt{ur}};
\node (c5) at (5.67, 3.51){\texttt{id}};
\node (c6) at (6.0, 2.81){\texttt{bn}};
\node (c7) at (7.02, 3.63){\texttt{hi}};

\end{tikzpicture}
}
\caption{The population of native speakers vs. the number of tokens collected by CulturaX, a representative multilingual corpus. \colorthr{Green marks} represent the four Chinese minority languages studied in this work, with \colortwo{orange marks} for several mid-resource languages in Asia and \colorfou{violet marks} for six languages with the richest resources. We obtain the population data from Wikipedia and check the credibility of the references. } % 设置caption
\label{fig:population_vs_token}  % 设置用于reference的label
\end{figure}
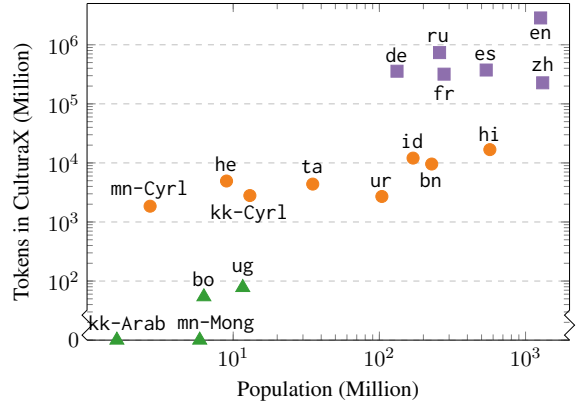

As illustrated in Figure~\ref{fig:population_vs_token}, the four languages, although spoken by tens of millions, face a critical deficiency in linguistic resources, posing obstacles to both academic research and the development of practical AI applications. 
Notably, there is no open-source corpus available for Kazakh in Arabic script and Mongolian in traditional Mongolian script -- two language variants in China adopting underrepresented writing systems.
In the case of Tibetan and Uyghur, although several multilingual datasets such as OSCAR~\cite{abadji-etal-2022-towards} and CulturaX~\cite{nguyen2023culturax} cover these languages, their quality falls short. 
Our preliminary study reveals a critical issue of language misidentification.
Up to 34\% of data in the Uyghur split of CulturaX is actually Kazakh or Arabic texts.
This raises concerns for future research in low-resource NLP, as conducting experiments on these significantly contaminated \textit{monolingual} corpora may yield misleading conclusions.

\begin{table*}[ht]
\small
\centering
\begin{tabular}{c|cc|c|c}
\toprule
\textbf{Name} & \textbf{ISO 639-1} & \textbf{ISO 639-3} & \textbf{Language Family} & \textbf{Writing System}  \\
\midrule
Tibetan & \texttt{bo} & \texttt{bod} & 	
Sino-Tibetan & Tibetan script \\
Uyghur & \texttt{ug} & \texttt{uig} & Turkic &  Uyghur Arabic script\\
Kazakh & \texttt{kk} & \texttt{kaz} & Turkic & Kazakh Arabic script \\
Mongolian & \texttt{mn} & \texttt{mvf} & Mongolic & Traditional Mongolian script \\
\bottomrule
\end{tabular}
\caption{ISO codes, languages families, and writing systems of the languages in MC$^\text{2}$.}
\label{tab:language_info}
\end{table*}

\begin{table*}[ht]
\small
\centering
\setlength\tabcolsep{5pt}
\begin{tabular}{l|ccccc|cc}
\toprule
 & \textbf{CC100} & \textbf{mC4} & \textbf{ROOTS} & \textbf{OSCAR} & \textbf{CulturaX} & \textbf{MC$^\text{2}$ (crawl)} & \textbf{MC$^\text{2}$ (full)} \\
\midrule
Tibetan & \xmark & \xmark & \xmark & 532M & 718M &  1.7G & 2.2G \\
Uyghur & 378M & \xmark & \xmark & 146M (\sout{220M})$^{\dag}$ & 338M (\sout{403M})$^{\dag}$ & 520M & 736M \\
Kazakh (Arabic) & \xmark & \xmark & \xmark & \xmark & \xmark & 937M & 937M \\
Mongolian (Traditional) & \xmark & \xmark & \xmark & \xmark & \xmark & 970M & 970M \\
\bottomrule
\end{tabular}
\caption{Dataset sizes of different multilingual datasets in the four minority languages. MC$^\text{2}$ (crawl) denotes the subset of our newly-collected web crawls.
MC$^\text{2}$ (full) is the complete set of MC$^\text{2}$, which additionally contains texts collected from existing resources.  
$^{\dag}$We observe many cases of language misidentification in the Uyghur split of OSCAR and CulturaX. We report the data sizes after manual re-identification. The crossed-out numbers in the brackets indicate the original data sizes.}
\label{tab:comparison_with_other_datasets}
\end{table*}

To facilitate more transparent and reproducible NLP research on the minority languages in China, we present MC$^\text{2}$, a \textbf{M}ultilingual \textbf{C}orpus of \textbf{M}inority Languages in \textbf{C}hina, which is the largest open-source corpus for these languages so far. 
During data collection, we adhere to a quality-centric principle. 
We carefully design strategies for the selection of web pages to crawl, ensuring the language purity of the crawled texts. 
Afterwards, we conduct a thorough data cleaning process, eliminating extraneous elements such as web page headers and retaining only meaningful content.
Our newly collected MC$^\text{2}$ dataset surpasses existing datasets by containing more than twice the amount of Tibetan and Uyghur data, while maintaining superior data quality.

MC$^{\text{2}}$ particularly emphasize underrepresented writing systems, recognizing their significance both technically and culturally. 
In languages with multiple writing systems, such as Kazakh and Mongolian, previous research has predominantly focused on the more common scripts~\cite{abadji-etal-2022-towards,nguyen2023culturax}.
We illustrate that transliterating texts from high-resource scripts into lower-resource ones may not be effective for training, especially when the script conversion is flawed. 
Moreover, the texts collected in different scripts inherently embody cultural nuances of various language communities. 
Through probing, we demonstrate that writing systems play a crucial role in developing culturally-aware NLP systems.

In addition, we continually train language models with MC$^{\text{2}}$. 
They achieve comparable performance to their counterparts trained with closed-source data.
This further validates the value of our data in facilitating transparent NLP research.

To summarize, we make the following contributions:
(1) We present MC$^{\text{2}}$, the largest open-source corpus to date for four underrepresented languages in China.
(2) We highlight quality issues, especially language misidentification, prevalent in previous corpora, which may threaten trustworthy research on low-resource languages.
(3) We reveal the cultural complexities arising from the multiplicity of writing systems within languages.

\section{Related Works}

\paragraph{Multilingual Corpus}
To facilitate the development of multilingual LLMs, a series of large-scale multilingual corpora has been released, including CC100~\cite{conneau-etal-2020-unsupervised}, mC4~\cite{raffel2020exploring}, ROOTS~\cite{laurenccon2022bigscience}, OSCAR~\cite{abadji-etal-2022-towards},  CulturaX~\cite{nguyen2023culturax}, and Madlad-400~\cite{kudugunta2024madlad}. 
These corpora cover both high-resource languages and a handful of low-resource ones.
Recently, there are also efforts in constructing open-source corpora for low-resource languages in certain regions~\cite{cahyawijaya-etal-2021-indonlg,teodorescu-etal-2022-cree,doddapaneni-etal-2023-towards}.
However, the minority languages in China are still underrepresented in existing datasets, which calls for more attention to constructing open-source corpus for these languages.

\paragraph{NLP for Minority Languages in China}
In terms of the minority languages in China, previous works try to improve the accessibility of these languages, by collecting a handful of annotated datasets for specific NLP tasks, each covering one or some languages investigated in our study. 
They mainly focus on three types of tasks: text classification~\cite{qun2017end,yang-etal-2022-cino,shi2022milmo}, question answering~\cite{sun2021teaching}, and machine translation~\cite{costa2022no,zhang2024teaching}.
The models trained specifically for the minority languages in China include CINO~\cite{yang-etal-2022-cino}, MiLMo~\cite{shi2022milmo}, and CMPT~\cite{li2022multi}.
However, none of these works releases their pre-training corpus.
In contrast, we release a large-scale high-quality corpus for four minority languages in China, which, we hope, will help relieve the data scarcity problem and improve transparency in research on low-resource languages.

\paragraph{Culturally-Aware NLP}
Cultural considerations are increasingly recognized as crucial in NLP research and applications~\cite{hershcovich-etal-2022-challenges}.
An important aspect is the connection between cultural factors and linguistic forms within languages.
Previous works on intra-language variations have primarily focused on dialects~\cite{zampieri2020natural,ziems-etal-2022-value,liu-etal-2023-dada} and sociolects~\cite{mccormack2011hexagonal,zhang-etal-2021-sociolectal}. 
However, the dimension of writing systems within languages has received little attention.
We demonstrate that texts in underrepresented writing systems may encode unique cultural nuances that differ significantly from those found in communities using more common scripts.

\section{Data Collection and Analysis}
Here we describe the procedure of creating MC$^{\text{2}}$, the largest open-source corpus so far for four minority languages in China.

Before collecting MC$^{\text{2}}$, we first conduct a preliminary audit in the low-resource language splits of previous multilingual corpora and find severe quality issues such as language contamination.
Therefore, we propose a quality-centric solution for data collection of low-resource languages, which aims to ensure accuracy while improving the comprehensiveness and coverage of the data. 

The collection procedure of MC$^{\text{2}}$ mainly consists of three steps: (1) gathering web crawls, (2) incorporating existing datasets, and (3) deduplicating and filtering.
Throughout the collection process, we adhere to the protocol of quality-centric collection.
We hope to establish a reliable groundwork for subsequent language model training or linguistic research.
Notably, our framework for data collection is generally language-agnostic and can be easily applied to collecting web corpora for other low-resource languages.

\subsection{Quality Issues in Previous Corpora}
\label{sec:quality_issues}
When auditing previous multilingual web corpora for low-resource languages~\cite{abadji-etal-2022-towards,nguyen2023culturax}, we find two critical quality issues: language misidentification and insufficient data cleaning. 
These defects pose a significant threat to effective model training and might undermine the credibility of research findings.

\paragraph{Language Misidentification} 
An important step in collecting web corpora for low-resource languages is to identify the language of a web page. 
However, current language identification tools such as fastText~\cite{joulin2016bag} are prone to error~\cite{kreutzer-etal-2022-quality}.
This issue is more severe in languages with similar writing systems.

Specifically, we audit the Uyghur splits in CulturaX and OSCAR. 
Since the documents in these two datasets are paired with their respective URLs, we can efficiently label their languages according to the website of their URLs.
We manually check the 654 websites comprising the Uyghur corpus in CulturaX. 
Our examination reveals that 16\% of the data is actually in Kazakh or Arabic, languages utilizing scripts akin to that of Uyghur.
Similarly, 33\% of the Uyghur corpus in OSCAR are Kazakh or Arabic. See the example of misidentification in Appendix~\ref{appendix:quality_issue_examples}.

This issue presents a significant concern for future research in low-resource NLP, as experiments conducted on heavily contaminated corpora can produce misleading results. 
For instance, this contamination can adversely impact research on language transfer. 
Researchers may aim to evaluate whether a model trained on a \textit{monolingual} corpus in Language A can transfer its capabilities to Language B. 
However, if the corpus already contains a considerable amount of Language B, the target language, the conclusions from such experiments are unlikely to be reliable.

\paragraph{Insufficient Data Cleaning} 
The web crawls in previous corpora often contain unwanted texts such as sidebars, headers, and footers, as shown the example in Appendix~\ref{appendix:quality_issue_examples}. 
We sample 100 pages from the Tibetan corpus of CulturaX and our manual annotation shows that 42\% contain headers or footers. 
These undesired texts affect the coherence of the document texts, which might hinder models from learning the linguistic patterns of low-resource languages.

\subsection{Step 1: Web Crawling}
\label{sec:web_crawling}
Our corpus is mainly made up of web crawls.
We combine both human labor and AI assistance to prevent the flaws in the previous corpora.

\paragraph{Language Identification} 
Different from previous efforts for curating \textit{massively-multilingual} web crawls~\cite{nguyen2023culturax}, the number of websites in each language is limited for the four minority languages in our study.
We manually maintain a list of high-quality websites for each language of our study, to avoid language contamination resulting from mislabeling by identification tools.
These websites are collected from a variety of sources, including URLs in existing corpora, search engines, web portals, website recommendations on social media, and hyperlinks on these websites. 
For each website in the list, we start from the homepage and crawl all the pages through a breadth-first search.

\paragraph{Text Extraction}
The web crawls in previous corpora often contain unwanted texts such as sidebars, headers, and footers. 
As the number of websites in our corpus is manageable and the web pages in the same website usually share the same structure, we automatically design text extraction rules for each website. 
These rules can precisely extract the title and main content of a web page, discarding other distracting texts. 
Specifically, for each website, we ask Github Copilot\footnote{\url{https://github.com/features/copilot}} to analyze the HTML structure of a sampled web page and write a Python code to extract its title and main content. 
In this way, we can extract the wanted texts in the raw web crawls with high accuracy and efficiency.

In total, we collected 2.0G Tibetan,  1.1G Uyghur, 1.3G Kazakh, and 1.7G Mongolian texts in our new web crawls.

\subsection{Step 2: Incorporation of Existing Datasets}
\label{sec:other_dataset}
Thanks to the existing efforts in the community, we incorporate open-source resources into our corpus, including \textbf{CulturaX}~\cite{nguyen2023culturax}, \textbf{Wikipedia}\footnote{\url{https://huggingface.co/datasets/graelo/wikipedia}}, and \textbf{NLGIW 2023 Shared Task}\footnote{\url{http://nlg.cipsc.org.cn/evaluation_12.html}}. 
We remove the data with language contamination when merging these datasets.

CulturaX is a cleaned collection of mC4 and OSCAR. 
Its Tibetan and Uyghur splits can be used to supplement our corpus.
However, there are errors in language identification in CulturaX, especially the confusion between Uyghur texts and Kazakh texts in Arabic scripts. 
We thus conduct language re-identification on the Uyghur split of CulturaX, using the website-centric strategy explained in Section~\ref{sec:quality_issues}. 
From CulturaX, we obtained 718M Tibetan, 338M Uyghur, and 65M Kazakh texts.

Wikipedia is a high-quality resource for low-resource languages. We collected 127M Tibetan and 41M Uyghur texts from Wikipedia. 
NLGIW 2023 Shared Task provides an open-source corpus of Tibetan news, where we add 342M texts.

\subsection{Step 3: Deduplication and Filtering}
Previous works claim that properly filtered and deduplicated web data is crucial to effective pretraining~\cite{lee-etal-2022-deduplicating,penedo2023refinedweb,ranathunga-etal-2024-quality}. 
Therefore, we take a series of measures for deduplication and filtering to ensure the high quality of our corpus.

\paragraph{Deduplication} 
We find two sources of duplication in our corpus.
One is that the websites in minority languages often repost from each other.
The other is the duplication between the web crawls in Step 1 and the existing corpus merged in Step 2. 
We thus use a combination of deduplication methods, including URL-based, exact, and fuzzy deduplication. 
See the implementation details in Appendix~\ref{appendix:dedup_filter}

\paragraph{Filtering} 

We design the filtering rules from several aspects, including repetition, document lengths, and unexpected characters. See details in Appendix~\ref{appendix:dedup_filter}.

We additionally make efforts to remove privacy information from the corpus. We use heuristics to identify emails, telephone numbers, and 
Resident Identity Card numbers. We replace them with special tokens, i.e., \texttt{[email]}, \texttt{[phone]}, and \texttt{[idcard]}.

After deduplication and filtering, we obtain 2.2G Tibetan, 736M Uyghur, 937M Kazakh, and 970M Mongolian texts, which compose the current version of MC$^\text{2}$.

\begin{figure}[t]
\centering
\includegraphics[scale=0.6]{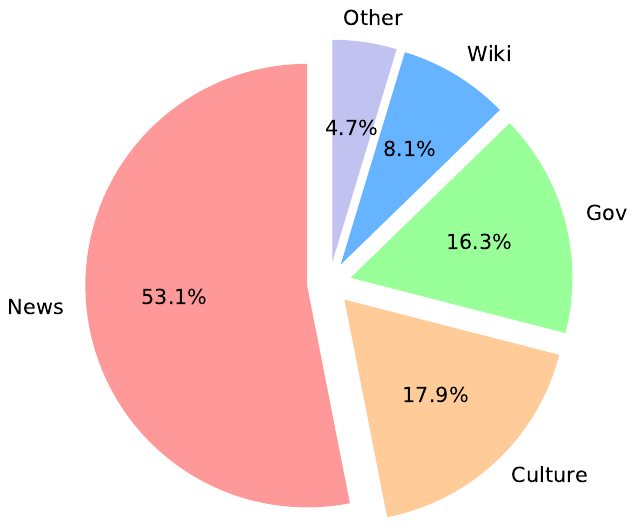}
\caption{Domain distributions of MC$^\text{2}$. The percentages are calculated from data sizes.}
\label{fig:domain}
\end{figure}

\subsection{Statistical Analysis of MC$^{\text{2}}$}

We provide an in-depth analysis from the perspective of domain diversity and document length.

\paragraph{Domain Diversity}
For the pretraining corpus, both quality and diversity are crucial. 
When curating the website list for crawling, we comprehensively collect high-quality websites from diverse domains to enhance the dataset's diversity while ensuring quality. 
Specifically, our data sources fall into five categories: News, Culture, Government, Wikipedia, and Others. 
Figure~\ref{fig:domain} shows the proportion of different categories of websites in MC$^\text{2}$.
Our corpus covers formal official documents, and miscellaneous news articles about politics, important events, daily life, and entertainment, as well as cultural content related to literature, art, and religion.
It is noteworthy that news articles constitute the majority of our datasets. 
News articles are usually grammatically correct, which could potentially facilitate learning language modeling under a low-resource setting.

\begin{figure}[t]
\centering
\includegraphics[scale=0.52]{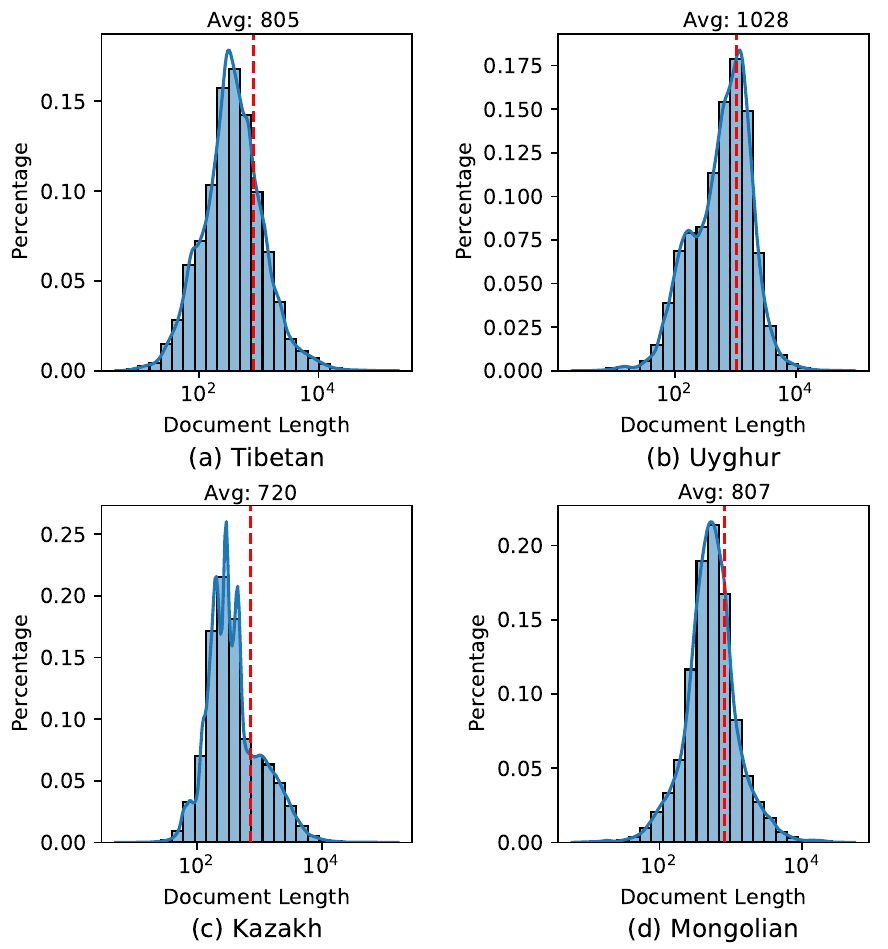}
\caption{Length distributions of the documents in MC$^\text{2}$ (using log scales for x-axes), calculated based on the tokens by the tokenizer of CINO~\cite{yang-etal-2022-cino}. }
\label{fig:length}
\end{figure}

\paragraph{Document Length}
Constrained by the absence of high-quality long-form texts, prior efforts like CINO~\cite{yang-etal-2022-cino} have to train their models with a collection of individual sentences, which might restrict their abilities on longer texts.
In contrast, our work introduces a document-level corpus.
We report the distribution of document length for each language in Figure~\ref{fig:length}. The average document length of MC$^\text{2}$ far exceeds the maximum input length, i.e., 512, of the previous BERT-size model for minority languages in China.

We observe that CultureX also includes long documents.
However, as mentioned in Section~\ref{sec:web_crawling}, its data cleaning might not be adequate, introducing undesired texts like sidebars, headers, and footers.
These issues could disrupt the semantic coherence of the entire document, making it challenging for models to learn correct dependencies within long texts.
We expect that our high-quality, long-form corpus will assist future efforts in training models with better capacities for understanding and generating longer texts in minority languages.

\begin{table}[t]
\small
\centering
\begin{tabular}{lccc}
\toprule
\textbf{Script} & \textbf{Region} & \textbf{Sample Text} & \textbf{Size} \\
\midrule
Arabic (\texttt{kk}) & China & \includegraphics[scale=0.5]{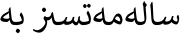} & 0  \\
Cyrillic (\texttt{kk}) & Kazakhstan &  \includegraphics[scale=0.5]{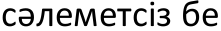} & 18G \\
\midrule
Traditional (\texttt{mn}) & China &
    \includegraphics[scale=0.56]{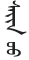}
     &  0 \\
Cyrillic (\texttt{mn}) & Mongolia & \includegraphics[scale=0.5,valign=b]{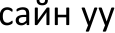} & 12G \\
\bottomrule
\end{tabular}
\caption{Comparison between the different writing systems of Kazakh (\texttt{kk}) and Mongolian (\texttt{mn}). The sample texts mean \textit{hello}. We report the data sizes in CulturaX.}
\label{tab:mn_kk_script_compare}
\end{table}

\section{Discussion on Underrepresented Writing Systems}
Many languages adopt distinct writing systems across various regions. For instance, in China, minority languages such as Kazakh and Mongolian employ scripts that differ from the Cyrillic scripts used in Kazakhstan and Mongolia, as illustrated in Table~\ref{tab:mn_kk_script_compare}.
Unfortunately, existing datasets predominantly concentrate on the more prevalent writing systems, neglecting the less common ones.
In response to this issue, MC$^{\text{2}}$ is the first effort to collect native corpora for the two underrepresented writing systems, i.e., the Kazakh Arabic script and the traditional Mongolian script.

To delve into the multiplicity of writing systems, we conduct empirical studies from both the \textbf{technical} and the \textbf{cultural} perspectives. 
We find that imperfect transliteration between writing systems harms model training. 
Furthermore, corpora in different writing systems of the same language encapsulate unique cultural context specific to their respective language communities.
These findings underscore the importance of gathering resources for minority writing systems.

\begin{figure}[t]
\centering
\includegraphics[scale=0.5]{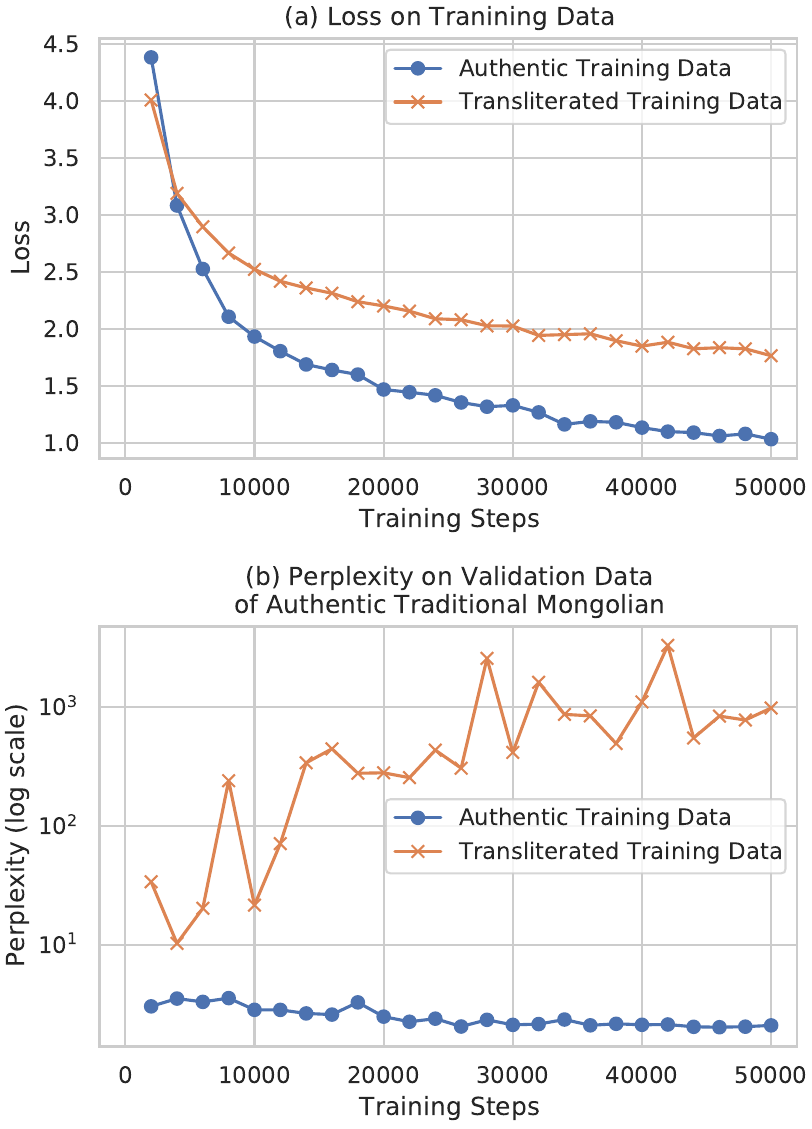}
\caption{Comparison between training XLM-RoBERTA-large with authentic and transliterated traditional Mongolian data. (a) The loss on the training data. (b) The perplexity on the hold-out evaluation data from the authentic traditional Mongolian corpus.}
\label{fig:mn_transliteration}
\end{figure}

\subsection{Flawed Training with Imperfect Transliteration}
A common approach to addressing different writing systems is transliteration~\cite{nakov-ng-2009-improved,chakravarthi2019comparison,muller-etal-2021-unseen,sun-etal-2022-alternative}. 
To obtain a model for low-resource scripts, it is intuitive to transliterate the corpus in the high-resource scripts into low-resource ones for training. 
However, there are no one-to-one conversion rules between scripts for languages such as Mongolian~\cite{feilong2014conversion}. 
The transliteration between traditional and Cyrillic Mongolian is context-dependent and current open-source tools are far from perfect. 
We investigate how the noisy transliteration results affect the training of models.

\paragraph{Experimental Setup} 
We prepare two versions of training data in traditional Mongolian.
One is 900M \textit{transliterated} traditional Mongolian data, which is converted from the Cyrillic Mongolian data in CulturaX, using the only open-source transliteration tool available so far\footnote{\url{https://github.com/tugstugi/mongolian-nlp}}. 
The transliterated results often contain errors, as shown by the examples in Appendix~\ref{appendix:mn_transliterate}.
The other version of data is an equivalent volume of \textit{authentic} traditional Mongolian data in MC$^{\text{2}}$.
We then train XLM-RoBERTa-large~\cite{conneau2019unsupervised} with these two different versions of data respectively and compare the convergence speed and performance of the models.

\paragraph{Results} 
The training processes are illustrated in Figure~\ref{fig:mn_transliteration}.
The training loss of transliterated data is significantly higher than that of authentic data. 
This discrepancy arises due to the substantial noise in transliterated data, making it challenging for the model to capture underlying linguistic patterns. 
We further measure the perplexity of both models on a hold-out validation set of traditional Mongolian text.
The perplexity of the model trained on authentic data steadily decreases with the training steps, maintaining a similar magnitude to the training loss. 
Conversely, the model trained on transliterated data exhibits a sharp increase in perplexity when evaluated on authentic data, indicating that it fails to learn the language modeling of true traditional Mongolian.

\paragraph{Lessons Learned} 
Using noisy data transliterated from high-resource writing systems will greatly hinder the learning of low-resource writing systems. 
This highlights the necessity of collecting an authentic corpus for underrepresented scripts in the absence of mature transliteration tools.

\subsection{Cultural Differences Behind Writing Systems}
For some languages such as Kazakh, we can achieve perfect transliteration between different writing systems using pre-defined rules.
Nevertheless, there exist disparities in the cultural backgrounds between the language variants using different scripts. 
With the technique of probing~\cite{jiang-etal-2020-x,zhang-etal-2021-sociolectal}, we investigate whether the training data collected from different writing systems will lead to distinct cultural knowledge in the resulting models.

\paragraph{Probing Questions}
We take the Kazakh language as our research target. 
The Kazakh community in China uses the Arabic script while the Cyrillic script is adopted in Kazakhstan.
We design probing questions that reflect the cultural differences between the two Kazakh communities, such as those in Table~\ref{tab:kk_probe}.
Each question offers two options: Option A aligns with the cultural context of the Kazakh community in China, while Option B corresponds to the cultural context of Kazakhstan.
These questions are manually formulated by comparing Wikipedia articles that describe the two communities.
We compile 25 pairs of questions covering topics such as geography, politics, and daily life.
Originally written in English, these questions are translated into Cyrillic Kazakh using Google Translate, and then transliterated into Arabic Kazakh.

\paragraph{Experimental Setup} 
We train two distinct Kazakh language models based on XLM-RoBERTa-large, each tailored to one of the writing systems. One is trained with 900M authentic Cyrillic Kazakh texts from CulturaX. And the other is trained with an equivalent volume of Arabic Kazakh texts from our MC$^{\text{2}}$ corpus. 
We subject the two models trained on different scripts to the cultural probing questions.
We query the Arabic Kazakh model with questions written in the Arabic script.
Similarly, for the Cyrillic Kazakh model, we use questions written in the Cyrillic script.

\paragraph{Results}
The Arabic Kazakh model selects Option A for 84\% of the probing questions, indicating a strong alignment with the cultural characteristics of Kazakh communities in China. Conversely, the Cyrillic Kazakh model chooses Option B for 56\% of the questions, reflecting the cultural characteristics of Kazakh communities in Kazakhstan.

The cultural distinction is probably inherent in the corpus collected from websites using different writing systems.
Subsequently, it is captured by the models trained on data in different scripts. 

In Table~\ref{tab:kk_probe}, we present three examples of probing questions. 
For Query 1, the two models yield divergent responses concerning the holiday celebrated on May 1st. This discrepancy arises from the fact that the official holidays on this date differ between China and Kazakhstan. Similarly, Query 2 and Query 3 reflect the cultural differences in terms of economy and geography. See their explanations in Appendix~\ref{app:explanation_probing}.

\paragraph{Lessons Learned} 
The corpora in different writing systems contain unique cultural context specific to their respective communities. 
Hence, it is sub-optimal to directly transliterate high-resource scripts into low-resource ones for training, which may undermine the cultural nuances underlying the low-resource writing system. 
The inclusiveness of writing systems is crucial to the construction of culturally-aware models.
We encourage future research on properly leveraging the data in the high-resource writing systems while preserving the cultural uniqueness in the underrepresented scripts.

\begin{table}[t]
\small
\centering
\begin{tabular}{lcc}
\toprule
\textbf{Option} & \textbf{Arabic} & \textbf{Cyrillic} \\
\midrule
\multicolumn{3}{l}{\textit{Query 1: On May 1st we celebrate \underline{\ \ \ \ \ \ \ \ }.}}\\
A. International Workers' Day & \textbf{2.48} & 5.46 \\
B. People's Unity Day & 3.23 & \textbf{3.30} \\
\midrule
\multicolumn{3}{l}{\textit{Query 2: The currency used here is \underline{\ \ \ \ \ \ \ \ }.}} \\
A. Renminbi & \textbf{1.97} & 7.29 \\
B. Kazakhstani Tenge & 3.33 & \textbf{6.55} \\
\midrule
\multicolumn{3}{l}{\textit{Query 3: There is a big lake near my home called \underline{\ \ \ \ \ \ \ \ }.}} \\
A. Sayram Lake & \textbf{3.33} & 3.15 \\
B. Lake Balkhash & 4.35 & \textbf{2.66} \\
\bottomrule
\end{tabular}
\caption{Knowledge probing on the two models trained with Kazakh data in different writing systems. The numbers are the perplexities of the corresponding options. }
\label{tab:kk_probe}
\end{table}

\begin{table*}[ht]
\small
\centering

\setlength\tabcolsep{4pt}
\begin{tabular}{l|c|ccccc|c}
\toprule
\multirow{2}{*}{\textbf{Model}} & \textbf{Open-source} & \multicolumn{5}{c|}{\textbf{Classification (Accuracy / Weighted F1)}} & \textbf{QA (EM / F1)} \\
&  \textbf{Corpus} & \texttt{bo} & \texttt{ug} & \texttt{kk} & \texttt{mn} & All & \texttt{bo} \\
\midrule
mBERT-base & No & 20.9 / 19.8 & 67.7 / 77.4 & 13.7 / 10.6 & 56.1 / 59.6 & 27.8 / 26.0 & \ \  3.6 /\ \ 4.6 \\
XLM-RoBERTa-large & No & 32.3 / 28.8 & 86.3 / 88.7 & 35.8 / 34.8 & 43.5 / 45.8 & 39.0 / 38.0 & \ \ 7.0 / 30.1 \\
CINO-large-v2 & No & \textbf{55.9} / \textbf{58.6} &  \textbf{88.0} / \underline{89.5} & \textbf{42.0} / \textbf{41.6} & \underline{62.4} / \underline{65.2} & \underline{50.4} / \textbf{50.1} & \underline{14.0} / \textbf{57.1} \\
MC$^{\text{2}}$XLMR-large (Ours) & Yes & \underline{52.6} / \underline{49.9} & \underline{87.7} / \textbf{89.7} & \underline{40.3} / \underline{37.4} & \textbf{72.6} / \textbf{74.4} & \textbf{52.0} / \underline{48.9} & \textbf{14.8} / \underline{53.6} \\

\bottomrule
\end{tabular}
\caption{Performance of different models under the zero-shot transfer setting. The best scores are made \textbf{bold}, with the second \underline{underlined}. For classification, we use accuracy and weighted F1 as metrics. For QA, we use EM and F1. }
\label{tab:main_experiment}
\end{table*}

\section{Continual Pretraining with MC$^{\text{2}}$}
To demonstrate the practical value of our corpus, we train models with MC$^{\text{2}}$ and compare their performance with competitive counterparts. 
Instead of training from scratch, we conduct continual pretraining on existing models with MC$^{\text{2}}$, which is an effective technique to adapt models to new languages~\cite{ebrahimi-kann-2021-adapt,muller-etal-2021-unseen,yong-etal-2023-bloom}.
We continually pretrain XLM-RoBerta-large~\cite{conneau2019unsupervised} with MC$^{\text{2}}$, obtaining \textbf{MC$^{\text{2}}$XLMR-large}.
Thanks to the high quality of MC$^{\text{2}}$, it performs comparably to CINO, a counterpart trained with closed-source data.
We also attempt to adapt larger models such as Llama~\cite{touvron2023Llama} to the four languages, obtaining \textbf{MC$^{\text{2}}$Llama-13B}. The newly trained model shows the superior ability of in-context learning in these languages.

\subsection{Model Training}
Two models with different architectures and sizes are used for continual pretraining.
One is XLM-RoBERTa-large~\cite{conneau2019unsupervised}, a 560M encoder-only multilingual model.
The other is Llama2-ZH-13B~\cite{Llama2-ZH}, a
bilingual model supporting both English and Chinese.

The vocabularies of XLM-RoBERTa-large and Llama2-ZH-13B hardly contain tokens for the four languages in our study. 
Thus, we add 3K new tokens for each language, obtained by BPE~\cite{sennrich-etal-2016-neural}. 

To avoid the catastrophic forgetting of learned languages, we add 0.25B tokens of Chinese data from Wanjuan~\cite{he2023wanjuan} and 0.25B tokens of English data from C4~\cite{raffel2020exploring} for training.
See more training details in Apppendix~\ref{appendix:model_training}.

\subsection{Evaluation Setup}
\paragraph{Datasets}
For the four minority languages, we can only find limited datasets for evaluation. 
\textbf{WCM-v2}~\cite{yang-etal-2022-cino} is a 10-category text classification dataset. 
\textbf{TibetanQA}~\cite{sun2021teaching} is a Tibetan machine reading comprehension (MRC) dataset. 
\textbf{Flores-200}~\cite{costa2022no} contains machine translation (MT) tasks for Tibetan and Uyghur. 
We additionally construct an MRC dataset, \textbf{XQuAD-MT}, by translating XQuAD~\cite{artetxe-etal-2020-cross} into Uyghur and Tibetan with Google Translate\footnote{\url{https://translate.google.com/}}. 
All the evaluated datasets only consist of testing sets.
See data statistics in Appendix~\ref{app:data}.

\paragraph{Settings}  
For smaller encoder-only models, we adopt \textbf{zero-shot transfer}, i.e., finetuning a model with instances in a high-resource language and directly testing it on the low-resource language instances~\cite{artetxe-etal-2020-cross}. 
For WCM-v2, we use its Chinese instances for training. 
For TibetanQA and XQuAD-MT, we use CMRC 2018~\cite{cui-etal-2019-span}, a Chinese MRC dataset for training.
For larger generative models, we adopt \textbf{in-context learning}, providing exemplars in the prompt~\cite{brown2020language}. 
See the prompts in Appendix~\ref{app:prompts}.

\paragraph{Compared Models} 
We compare our models with a wide range of multilingual models, including CINO~\cite{yang-etal-2022-cino}, mBERT~\cite{devlin-etal-2019-bert}, XLM-RoBERTa~\cite{conneau2019unsupervised}, BLOOM~\cite{workshop2022bloom}, mT5~\cite{xue-etal-2021-mt5} and ByT5~\cite{xue-etal-2022-byt5}.
These models (potentially) process one or more languages in our study. 
XLM-RoBERTa is a multilingual pretrained language model (PLM), whose vocabulary contains 3 Tibetan tokens, 5 Mongolian tokens, and more than 14K tokens in the Arabic script. However, it has only been pretrained in Uyghur. 
CINO is the largest PLM so far focusing on the minority languages in China, but its pretraining corpora are closed source. 
BLOOM, mT5, and ByT5 are multilingual LLMs, which can represent tens of languages. 
Although they are not trained in the four minority languages, their tokenizers can encode unseen languages instead of treating them as unknown tokens.  
See more details of the supported languages and numbers of parameters in Appendix~\ref{app:prompts}.

\subsection{Results}
We show the experiment results of zero-shot transfer in Table~\ref{tab:main_experiment}, with XLM-RoBERTa-large as the baseline.
We find both MC$^{\text{2}}$XLMR-large and CINO can outperform vanilla XLM-RoBERTa-large on the classification and MRC tasks. It indicates that pretraining on minority language corpora can effectively enhance the model's ability to represent these languages. 
When comparing the two models for Chinese minority languages, we find our MC$^{\text{2}}$XLMR-large can exhibit comparable performance to CINO, which is trained on a closed-source corpus three times larger than MC$^2$. 
This proves the high quality of MC$^{\text{2}}$, which can contribute to more transparent and reproducible research on minority languages in China.

\begin{table}[t]
\small
\centering

\begin{tabular}{lcc}
\toprule
\multirow{2}{*}{\textbf{Model}} & \textbf{WCM-v2} & \textbf{TibetanQA} \\
 & Acc. / W.F1 & EM / F1\\
\midrule

BLOOM-7.1B & \ \ 9.2 / \ \ 8.9 & \ \ \ 0.0 / \ \ 1.3 \\
mT5-xxl  & 10.0 / 10.4 & \ \ \ 0.0 / \ \ 2.6 \\ 
ByT5-xxl & 	10.2 / 10.6 &  \ \ \ 2.5 / \textbf{33.1} \\
MC$^{\text{2}}$Llama-13B (Ours) & \textbf{36.9} / \textbf{37.0} & \ \ \textbf{\ 3.8} / 31.5 \\
\bottomrule
\end{tabular}
\caption{Performance of LLMs under in-context learning. For WCM-v2, we report averaged accuracy and weighted F1 of four languages. For TibetanQA, we use EM and F1 as metrics.}
\label{tab:llm_experiment}
\end{table}

Table~\ref{tab:llm_experiment} illustrates the results of LLMs under the in-context learning setting. 
The performance of evaluated multilingual LLMs on WCM-v2 resembles random guessing, with an accuracy of around 10\% in the ten-class classification tasks. 
These models struggle with the Tibetan MRC task. 
In contrast, the MC$^{\text{2}}$Llama-13B model, pretrained on the MC$^{\text{2}}$ corpus, outperforms others by more than +26\% accuracy and F1 on WCM-v2. It also achieves 31.5\% F1 on TibetanQA.
These results show that MC$^{\text{2}}$Llama-13B has a superior ability to handle tasks in the four minority languages of China. 
Besides smaller models, MC$^{\text{2}}$ can also be leveraged for the pretraining of LLMs, thereby effectively enhancing the models' ability to represent these languages.

The results on Flores-200 and XQuAD-MT are reported in Appendix~\ref{app:additional_exp}. 
The models trained on MC$^2$ also demonstrate superior performance on these tasks.

\section{Conclusion}
We present MC$^{\text{2}}$, an open-source corpus for four minority languages in China.
It is also the first corpus focusing on two underrepresented writing systems, i.e., the Kazakh Arabic script and the traditional Mongolian script.
To address the severe quality issues in previous low-resource datasets, we adopt a quality-centric strategy to collect MC$^{\text{2}}$.
We also emphasize the cultural significance of including low-resource writing systems through empirical studies.
Our data and models are openly accessible to the community to facilitate research and applications on low-resource languages.

\section*{Limitations}
\paragraph{Data Sources}
MC$^{\text{2}}$ is mainly composed of web crawls. We have done our best to enrich the domains covered by the MC corpus. However, it is not feasible to collect data from all websites worldwide. In this work, due to copyright constraints, we are also unable to collect and publicly release a corpus sourced from books in these minority languages. We note that the texts from books can also be used to pretrain language models.

News articles are one of our primary data
sources, which are crawled from public websites. The data could potentially contain the biases of these media. Therefore, the data should be used with caution to avoid potential biases and misrepresentations.

\paragraph{Probing Questions}
It is challenging to collect the kind of probing question in our study, which requires mining the differences between the two communities and formulating them as multi-choice questions.
We made great efforts and collected 25 verified probing questions covering a diverse range of topics.
As a preliminary study, the size of probing questions is small. 
We plan to design efficient methods for cultural question sourcing and expand the data scale. 

\section*{Acknowledgments}
This work is supported in part by NSFC
(62161160339) and Beijing Science and Technology Program (Z231100007423011).
We thank the anonymous reviewers for their valuable suggestions.
We thank native speakers of minority languages who contribute to this work.
For any correspondence, please contact Yansong Feng.

% Entries for the entire Anthology, followed by custom entries
\bibliography{anthology,custom}

\clearpage
\appendix

\section{Corpus Details}
\subsection{Quality Issues in Previous Corpora}
\label{appendix:quality_issue_examples}
We provide several cases of quality issues in previous corpora.

\paragraph{Language Misidentification} 
In the Uyghur split of CulturaX, the website \texttt{kazakh.people.com.cn} ranks third in terms of document count, comprising 9\% of the total documents in the dataset.
However, this website is a Kazakh-language news website.
These documents are mislabeled as Uyghur because the website uses the Kazakh Arabic script, which is similar to the Uyghur Arabic script.

\begin{figure}[ht]
\centering
\includegraphics[scale=0.45]{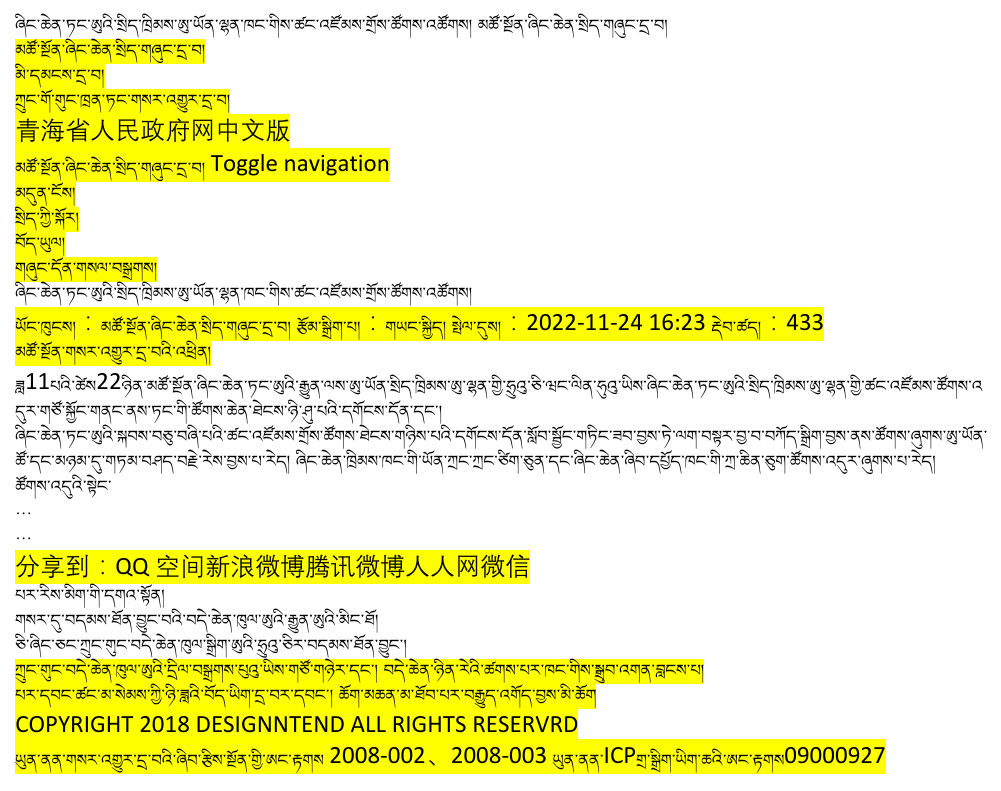}
\caption{An example of insufficient data cleaning from the Tibetan corpus in CulturaX. The highlighted parts are irrelevant texts on a web page, such as headers and footers.}
\label{fig:header_footer}
\end{figure}

\paragraph{Insufficient Data Cleaning}
In Figure~\ref{fig:header_footer}, we show an example of insufficient data cleaning from the Tibetan corpus in CulturaX. The highlighted parts are irrelevant texts on a web page, such as headers and footers.

\subsection{Deduplication and Filtering}
\label{appendix:dedup_filter}
We describe in detail the deuplication and filtering process of MC$^{\text{2}}$.

\paragraph{Deduplication}
We first discard the web pages from different sources with the same URLs. 
We then adopt both exact and fuzzy deduplication. For exact deduplication, we use the SHA-256 hash function to calculate the hash value of each web page's content and keep only one page for the pages with the same hash value. 
For fuzzy deduplication, we remove similar web pages by applying MinHash~\cite{666900}: for each page, we compute the minhash values and measure its approximate similarity with other pages, removing pairs whose minhash values are the same in at least one bucket. 
We use the same parameters as \citet{lee-etal-2022-deduplicating}: $n = 5$ (5-grams), $b = 450$, and $r = 20$.

\paragraph{Filtering}
We use the following rules:
\begin{itemize}
    \item Repetition: We remove the documents that have a high ratio of paragraph repetition or n-gram repetition.
    \item Document Length: We remove the documents whose lengths are below the threshold.
    \item Unexpected Character: We remove the document whose ratios of the characters in the target language are below the threshold.
\end{itemize}
Note that the hyperparameters in the heuristic for filtering English corpora are not necessarily applicable to low-resource languages due to the difference in encodings and linguistic characteristics. One needs to decide the hyperparameters for each language respectively.

\subsection{Code-Switching in MC$^2$}
Code-switching is an important and interesting phenomenon in multilingual NLP. 
Here we share some analyses and findings of the code-switching phenomenon present in our corpus.
We mainly focus on the code-switching between Chinese and the four minority languages in China.

First, we calculate the proportion of documents containing Chinese characters for the four languages in MC$2$. The results are shown in Table~\ref{tab:code-mixing_chinese-char}.
3.3\% of the documents in MC$^2$ contain Chinese characters, indicating the commonness of the code-switching phenomenon in our corpus.
Uyghur exhibits the highest rate of code-switching. 
We find that the Uyghur documents in our corpus are more likely to attach Chinese translations after entity mentions.

\begin{table}[t]
\small
\centering
\setlength\tabcolsep{3pt}
\begin{tabular}{lr}
\toprule
\textbf{Subset} & \textbf{\%}  \\
\midrule
Tibetan	& 2.5\% \\
Uyghur &	9.3\% \\
Kazakh &	0.9\% \\
Mongolian &	0.7\% \\
\bottomrule
\end{tabular}
\caption{Percentage of documents containing Chinese characters for the four subsets in MC$^2$.}
\label{tab:code-mixing_chinese-char}
\end{table}

Second, we analyze the types of code-switching present in our corpus. Inspired by previous works~\cite{dogruoz-etal-2021-survey,winata-etal-2023-decades}, we classify documents containing Chinese characters into three types:
\begin{itemize}
    \item Intra-sentential code-switching: switching that occurs in a sentence where a word was switched to another language. 
    \item Inter-sentential code-switching: switching that occurs in two different sentences.
    \item Attached translations of entities: In many cases, authors of articles often append Chinese translations after entity mentions. Strictly speaking, this doesn't fall under code-switching, but this situation is quite common.
\end{itemize}

We sample 50 documents containing Chinese characters and manually classify them into three types. The proportion of each type is shown in Table~\ref{tab:code-mixing_type}.
The most common phenomenon is attaching Chinese translations to the entities mentioned in the documents. The common entities under these circumstances include scientific and cultural terms.
Regarding intra-sentential code-switching, the switched phrases are often entities of Chinese origin, such as Han Chinese names and literary works written in Chinese. These Chinese terms currently lack official translations in these languages, so they are directly borrowed in their original form.
Many instances of inter-sentential code-switching are sentences in different languages, where the sentences before and after each other are translations of one another. 
Based on this property, we can mine parallel sentences from the corpus, which we leave as future work.

\begin{table}[t]
\small
\centering
\setlength\tabcolsep{3pt}
\begin{tabular}{lrrrr}
\toprule
\textbf{Type} & \textbf{\texttt{bo}} & \textbf{\texttt{ug}} & \textbf{\texttt{kk}} & \textbf{\texttt{mn}}    \\
\midrule
Intra-sentential	& 6\% & 14\% & 38\% & 36\% \\
Inter-sentential &	46\%  & 0\% & 28\% & 24\% \\
Attached translations &	48\% & 86\% & 34\% & 40\% \\
\bottomrule
\end{tabular}
\caption{Percentage of documents containing Chinese characters for the four subsets in MC$^2$.}
\label{tab:code-mixing_type}
\end{table}

\section{Cultural Probing}

\label{app:explanation_probing}
We provide a detailed explanation of the probing cases in Table~\ref{tab:kk_probe}.
For Query 1, May 1st is a public holiday in both China and Kazakhstan. However, the people in the two countries celebrate different festivals. In China, people celebrate International Workers’ Day on May 1st, to commemorate the struggles of laborers and the working class for the eight-hour workday. However, according to the Kazakhstan law signed on October 18, 1995, May 1st was renamed People's Unity Day, and the Soviet-era Labor Day was formally canceled.

For Query 2, in mainland China, the official currency is Renminbi. Meanwhile, in Kazakhstan, the official currency is Kazakhstani Tenge.

For Query 3, Sayram Lake is an endorheic freshwater lake in the northern Tianshan Mountains, near to Ili Kazakh Autonomous Prefecture in China. Its name, \textit{Sayram}, is believed to originally derive from Kazakh, meaning \textit{blessing}. Sayram Lake is famous in China as \textit{the last drop of the Atlantic tears}. Lake Balkhash is also famous in Kazakhstan since it is the largest lake in Kazakhstan. Around 20\% of Kazakhstani people are living in the drainage basin of Lake Balkhash, including the residents of Almaty, the largest city in Kazakhstan.

\section{Mongolian Transliteration}
\label{appendix:mn_transliterate}
\begin{figure}[ht]
\centering
\includegraphics[scale=0.5]{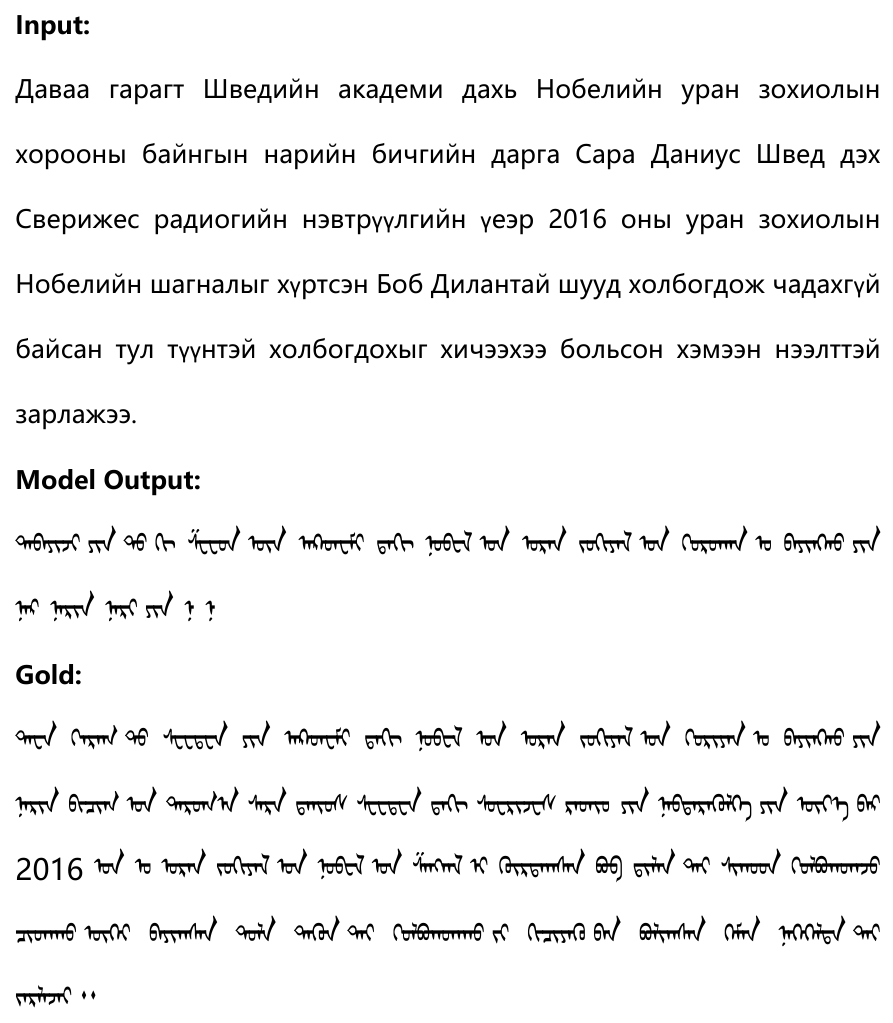}
\caption{An example of low-quality transliteration from Cyrillic Mongolian to traditional Mongolian by an open-source tool.}
\label{fig:bad_mn_transliterate}
\end{figure}

We use an open-source tool to transliterate Cyrillic Mongolian into traditional Mongolian\footnote{\url{https://github.com/tugstugi/mongolian-nlp/tree/master/bichig2cyrillic}}.
It is trained on 80K sentence pairs crawled from a closed-source transliteration system.
When transliterating a Cyrillic Mongolian corpus, we find that the transliteration quality deteriorates with the increase in the input length.
So we split each document into sentences and do the transliteration at a sentence level to improve the quality.
However, we find that the transliteration results are imperfect, especially in the case of long sentences.
In Figure~\ref{fig:bad_mn_transliterate}, the transliteration tool produces a bad result for a long input sentence, repeating similar words at the end of the output. 

\section{Experiment Details}
\subsection{Model Training}
\label{appendix:model_training}
\paragraph{Hyperparameters}
We list in Table~\ref{tab:hyperparameters} the hyperparameters for the continual pretraining of MC$^{\text{2}}$XLMR-large and MC$^{\text{2}}$Llama-13B.

\paragraph{Training Frameworks}
For MC$^{\text{2}}$XLMR-large, we use Transformers~\cite{wolf-etal-2020-transformers} for training. For MC$^{\text{2}}$Llama-13B, we use Megatron~\cite{shoeybi2019megatron} for training.

\paragraph{Computing Infrastructure}
We continually pretrain MC$^{\text{2}}$XLMR-large on one A100 GPU, and an epoch takes 17 hours.
We continually pretrain MC$^{\text{2}}$Llama-13B on eight A100 GPUs, and an epoch takes 20 hours.

\begin{table}[t]
\small
\centering
\setlength\tabcolsep{3pt}
\begin{tabular}{lcc}
\toprule
\textbf{Hyperparameter} & \textbf{MC$^{\text{2}}$XLMR-large} &  \textbf{MC$^{\text{2}}$Llama-13B} \\
\midrule
Max. seq. length & 512  & 2,048\\
Batch size (tokens) & 16K & 1M \\
Learning rate & 1e-4 & 2e-5\\
Warmup step & 5K & 200 \\
Epoch & 3 & 1 \\
\bottomrule
\end{tabular}
\caption{Hyperparameters for training MC$^{\text{2}}$XLMR-large and MC$^{\text{2}}$Llama-13B.}
\label{tab:hyperparameters}
\end{table}

\subsection{Data Statistics}
\label{app:data}
\paragraph{WCM-v2} 
Wiki-Chinese-Minority-v2(WCM-v2) is a classification dataset. It covers Mongolian, Tibetan, Uyghur, Cantonese, Korean, Kazakh and Chinese. However, the training set is only available in Chinese. 
The data statistics is shown in Table~\ref{tab:WCM_data_stats}

\begin{table}
\small
    \centering
    \setlength\tabcolsep{5pt}
    \begin{tabular}{l|rrrrr}
      \toprule
      \textbf{Category} & \texttt{mn} & \texttt{bo} & \texttt{ug} & \texttt{kk} & \texttt{zh-Train} \\%& zh-Dev & zh-Test \\
      \midrule
      Art & 135 & 141 & 3 & 348 & 2,657 \\%& 331 & 335 \\
      Geography & 76 & 339 & 256 & 572 & 12,854 \\%& 1589 & 1644 \\
      History & 66 & 111 & 0 & 491 & 1,771 \\%& 227 & 248 \\
      Nature & 7 & 0 & 7 & 361 & 1,105 \\%& 134 & 110\\
      Natural Science & 779 & 133 & 20 & 880 & 2,314 \\%& 317 & 287\\
      People & 1,402 & 111 & 0 & 169 & 7,706 \\%& 953 & 924\\
      Technology & 191 & 163 & 8 & 515 & 1,184 \\%& 134 & 152\\
      Education & 6 & 1 & 0 & 1,392 & 936 \\%& 130 & 118\\
      Economy & 205 & 0 & 0 & 637 & 922 \\%& 113 & 109\\
      Health & 106 & 111 & 6 & 893 & 551 \\%& 67 & 73\\
      \midrule
      Total & 2,973 & 1,110 & 300 & 6,258 & 32,000 \\%& 3995 & 4000\\
      \bottomrule
    \end{tabular}
    \caption{Number of instances in \textbf{WCM-v2}. Besides Chinese(zh), all minority languages only have test sets.}
    \label{tab:WCM_data_stats}
\end{table}

% \footnote{\url{http://cmli-nlp.com/}}
\paragraph{TibetanQA} Tibetan Machine Reading Comprehension Dataset (TibetanQA) is a Tibetan reading comprehension dataset, which contains 20,000 question-answer pairs and 1,513 articles. However, the released dataset contains only 2007 question-answer pairs. The average lengths (in characters) of contexts, questions, and answers are 688, 58, and 93 respectively. 

\paragraph{Flores-200} Flore-200 is a machine translation dataset for low-resource languages, which covers two languages of our study, i.e., Tibetan and Uyghur. We sample 200 testing instances for Tibetan-to-Chinese and Uyghur-to-Chinese translations.

\paragraph{XQuAD-MT} 
Since there are no existing QA datasets for the languages of interest except Tibetan, we translate the Chinese split of XQuAD into Uyghur and Kazakh through Google Translate\footnote{Google Translate does not support traditional Mongolian currently. As the Kazakh translation results by Google Translate are in the Cyrillic script, we transliterate them into the Kazakh Arabic script.}. 
We translate an instance's passage, question, and answer separately. Consequently, the translated answer may not exactly appear in the translated passage. 
We then extract an EXACT subset from the translated dataset where the translated answer exactly appears in the translated passage.
We obtain 1,190 QA pairs for Uyghur, 257 of which belong to the EXACT subset, and 1,190 QA pairs for Kazakh, 417 of which belong to the EXACT subset.

\subsection{Evaluation Setup}
\label{app:prompts}

\paragraph{Models}
We list the numbers of parameters and supported languages of all models we use in Table~\ref{tab:model_info}. We only consider four minority languages used in our experiments. Other supported languages can be found in these models' original papers.

\paragraph{Prompts}
Regarding in-context learning, we follow the previous work \cite{shi2022language} to use native exemplars with a high-resource language (Chinese) for prompt description. We use 3 exemplars for WCM-v2, 2 exemplars for TibetanQA, and 5 exemplars for Flores-200. 
In Table~\ref{tab:prompts}, we list the format of our prompts \textit{along with English translation in italics}.

\paragraph{Result Reporting}
We report the results of single runs. In the experiments of in-context learning, the models sometimes fail to output a label in the label list. In these cases, we assign random labels to these samples.

\begin{table*}[ht]
\small
\centering
\setlength\tabcolsep{4pt}
\begin{tabular}{l|c|cccc}
\toprule
\multirow{2}{*}{\textbf{Model}} & \multirow{2}{*}{\textbf{Params}} & \multicolumn{4}{c}{\textbf{Supported Languages}}  \\
&  & \textbf{Tibetan} & \textbf{Uyghur} & \textbf{Kazakh (Arabic)} & \textbf{Mongolian (Traditional)} \\
\midrule
mBERT-base & 110M & \xmark & \xmark & \xmark & \xmark \\
XLM-RoBERTa-large & 560M & \xmark & \checkmark & \xmark & \xmark \\
CINO-large-v2 & 442M & \checkmark & \checkmark & \checkmark & \checkmark \\
BLOOM-7.1B & 7.1B & \xmark & \xmark & \xmark & \xmark\\
mT5-xxl & 13B & \xmark & \xmark & \xmark & \xmark \\
ByT5-xxl & 13B & \xmark & \xmark & \xmark & \xmark \\
MC$^{\text{2}}$XLMR-large (Ours) & 572M & \checkmark & \checkmark & \checkmark & \checkmark \\
MC$^{\text{2}}$Llama-13B (Ours) & 13B & \checkmark & \checkmark & \checkmark & \checkmark \\
\bottomrule
\end{tabular}
\caption{The supported languages of the models in our experiments and their parameters.}
\label{tab:model_info}
\end{table*}

\begin{table*}
\centering
\small
\setlength{\tabcolsep}{5pt}
\begin{tabular}{p{2.0\columnwidth}r}
\toprule
\textbf{WCM-v2} (Classification) \\
\zhsmall{请仿照示例，将下面的\{藏语/维吾尔语/哈萨克语/蒙古语\}文本分类为艺术、地理、历史、自然、自然科学、人物、技术、教育、经济、健康中的一个类别。} \\
\textit{Follow the example and classify the following \{Tibetan/Uyghur/Kazakh/Mongolian\} passage into one of the categories of Art, Geography, History, Nature, Natural Science, People, Technology, Education, Economy, and Health.}\\
\zhsmall{示例\{1/2/3\}} \qquad \textit{Example \{1/2/3\}}\\
\zhsmall{文本：\texttt{\{Passage written in Tibetan\}}} \qquad \textit{Passage: \texttt{\{Passage written in Tibetan\}}}\\
\zhsmall{类别：\texttt{\{Class\}}} \qquad \textit{Class: \texttt{\{Class\}}}\\
\zhsmall{请仿照以上示例，将下面的\{藏语/维吾尔语/哈萨克语/蒙古语\}文本分类为艺术、地理、历史、自然、自然科学、人物、技术、教育、经济、健康中的一个类别。}\\
\textit{Follow the example above to classify the following \{Tibetan/Uyghur/Kazakh/Mongolian\} passage into one of the categories of Art, Geography, History, Nature, Natural Science, People, Technology, Education, Economy, and Health.}\\

\midrule
\textbf{TibetanQA} (Question Answering) \\
\zhsmall{阅读文章，从文章中抽取内容回答问题。文章、问题和答案都应是藏文。接下来是两个样例。}\\
\textit{Read the passage and answer the questions by extracting content from the passage. Passages, questions, and answers should be written in Tibetan. Here are two examples.}\\
\zhsmall{样例\{1/2\}} \qquad \textit{Example \{1/2\}}\\
\zhsmall{文章: \texttt{\{Context written in Tibetan\}}} \qquad \textit{Passage: \texttt{\{Context written in Tibetan\}}}\\
\zhsmall{问题: \texttt{\{Question written in Tibetan\}}} \qquad \textit{Question: \texttt{\{Question written in Tibetan\}}}\\
\zhsmall{答案: \texttt{\{Answer written in Tibetan\}}} \qquad \textit{Answer: \texttt{\{Answer written in Tibetan\}}}\\
\bottomrule
\end{tabular}
\caption{Prompt templates used in our experiments. The \textit{text in italics} are the English translations of the Chinese instructions.}
\label{tab:prompts}
\end{table*}

\section{Additional Results}
\label{app:additional_exp}

\subsection{WCM-v2}
In Table~\ref{tab:all_llm_exp}, we report the performance in different languages on WCM-v2.

\begin{table*}[!t]
\small
\centering
\setlength\tabcolsep{4pt}
\begin{tabular}{l|c|ccccc}
\toprule
\multirow{2}{*}{\textbf{Model}} & \textbf{Open-source} & \multicolumn{5}{c}{\textbf{Classification (Accuracy / Weighted F1)}} \\
&  \textbf{Corpus} & \texttt{bo} & \texttt{ug} & \texttt{kk} & \texttt{mn} & All  \\
\midrule

BLOOM-7.1B & Yes & 11.8 / \ 9.5 & \ \ 5.7 / \ \ 8.9 &  \ \ 8.4 / \ \ 7.4 & \ \ 9.9 / 13.0 & \ \ 9.2 / \ \ 8.9  \\
mT5-xxl  & Yes & \ \ 9.2 / 11.7 & 13.3 / 20.1 &  10.6 / 11.3 &  10.1 / 13.6 & 10.0 / 10.4 \\ 
ByT5-xxl  & Yes & 10.7 / 12.7 & \ \ 8.6 / 14.7 &  10.3 / 10.9 &  \ \ 9.9 / 12.9 & 10.2 / 10.6 \\ 
MC$^{\text{2}}$Llama-13B (Ours) & Yes & \textbf{30.7} / \textbf{33.7} & \textbf{39.0} / \textbf{52.5} & \textbf{34.0} / \textbf{32.4} & \textbf{45.4} / \textbf{51.9} & \textbf{36.9} / \textbf{37.0}  \\
\bottomrule
\end{tabular}
\caption{Performance of different models on WCM-v2 under in-context learning. The best scores are made \textbf{bold}. We use accuracy and weighted F1 as metrics. }
\label{tab:all_llm_exp}
\end{table*}

\subsection{Flores-200}
Considering that encoder-only models cannot perform the MT task, we use generative models for experiments. Since the models have limited abilities to generate texts in low-resource languages, we only conduct experiments to translate the low-resource languages into Chinese.

The results are shown in Table~\ref{tab:flores}. 
As they are not finetuned for the translation task, the four models generally perform poorly on this task merely through in-context learning. However, our model achieves non-zero scores, demonstrating its preliminary ability to understand these languages, thanks to the training on MC$^2$. We leave the supervised training setting of MT for future work.

\begin{table*}[t]
\small
\centering
\begin{tabular}{l|cc}
\toprule
\textbf{Model} &  \textbf{Tibetan} & \textbf{Uyghur} \\
\midrule
BLOOM-7.1B & 0.4 / 1.9 & 0.9 / 3.4\\
mT5-xxl  & 0.0 / 0.3 & 0.0 / 0.4 \\ 
MC$^{\text{2}}$Llama-13B (Ours) & \textbf{2.5} / \textbf{4.4} & \textbf{6.5} / \textbf{8.2} \\
\bottomrule
\end{tabular}
\caption{Performance of different models on Flores-200 under in-context learning. The best scores are made \textbf{bold}. The scores in each cell are BLEU and chrF. }
\label{tab:flores}
\end{table*}

\subsection{XQuAD-MT}

In Table~\ref{tab:xquad-mt} we report the performance on the full set and EXACT subset of XQuAD-MT.

\begin{table*}[!t]
\small
\centering
\begin{tabular}{l|cc|cc}
\toprule
\multirow{2}{*}{\textbf{Model}} & \multicolumn{2}{c|}{\textbf{Uyghur}} & \multicolumn{2}{c}{\textbf{Kazakh}} \\
&  Full & EXACT & Full &  EXACT \\
\midrule
mBERT-base & 0.2 / 3.8 &  \ \ 0.8 /\ \ 6.2 & 0.3 / \ \ 6.0 & 1.0 / \ \ 9.6 \\
XLM-RoBERTa-large & \textbf{2.4} / 9.5 & \textbf{10.9} / \textbf{20.9} & 2.2 / 13.5 & \underline{6.5} / 24.0 \\
CINO-large-v2 & 	1.8 / \textbf{9.8} & \ \ 8.6 / \underline{20.7} & \textbf{2.7} / \underline{13.9} & \textbf{7.7} / \underline{24.9} \\
MC$^2$XLMR-large (Ours) & \underline{2.0} / \underline{9.6} & \ \  \underline{9.3} / 19.5 & 	\underline{2.3} / \textbf{16.4} & \underline{6.5} / \textbf{27.0} \\
\bottomrule
\end{tabular}
\caption{Performance of different models on XQuAD-MT under zero-shot transfer. The best scores are made \textbf{bold}, with the second \underline{underlined}. The scores in each cell are EM and F1. }
\label{tab:xquad-mt}
\end{table*}

\end{document}